\documentclass[letterpaper, 10 pt, conference]{ieeeconf}  % Comment this line out if you need a4paper

\IEEEoverridecommandlockouts                              % This command is only needed if 
\usepackage{graphicx}
\usepackage[dvipsnames]{xcolor}

\usepackage{amsmath}
\usepackage{amssymb}
\usepackage{amsfonts}
\usepackage{mathtools}
\usepackage{mathrsfs}

\usepackage{multirow}
\usepackage{booktabs, tabularx, makecell, colortbl, array}
\newcolumntype{Y}{>{\centering\arraybackslash}X}
\usepackage{siunitx}
\usepackage{vcell}

\usepackage{tikz}
\usepackage{wrapfig}
\usepackage{float}
\usepackage[font=footnotesize]{caption}
\usepackage{subfigure}
\usepackage{svg}

\usepackage{tcolorbox}
\usepackage{listings}
\usepackage{courier}

\usepackage{xspace}

\usepackage{amsthm}

\usepackage{algorithm}
\usepackage[noend]{algpseudocode}

\let\labelindent\relax
\usepackage{enumitem}

\usepackage{url}
\usepackage{soul}
\usepackage{upgreek}
\usepackage{mwe}
\usepackage{calc}
\usepackage{pifont}
\usepackage{marvosym}
\usepackage{cite}

\usepackage{hyperref}

\definecolor{DarkGreen}{RGB}{0,128,0}
\definecolor{mymagenta}{HTML}{E1208A}
\title{\LARGE \bf
RATE: Risk-Aware Tactile Encoding for Contact-rich Robotic Manipulation}

\author{Yuyao Jiang, Haichao Liu, Jiarui Zheng, Zihan Ding,
Weihao Yuan, and Ziwei Wang
\thanks{Yuyao Jiang, Haichao Liu, Jiarui Zheng, Zihan Ding, and
Ziwei Wang are with the School of Electrical and Electronic Engineering,
Nanyang Technological University, Singapore (corresponding author:
Ziwei Wang; e-mails: yjiang062@e.ntu.edu.sg; haichao.liu@ntu.edu.sg;
jzheng025@ntu.edu.sg; zihan.ding@ntu.edu.sg; ziwei.wang@ntu.edu.sg).}
\thanks{Weihao Yuan is with Daimon Robotics, Shenzhen, China
(e-mail: dr.weihao.yuan@gmail.com).}
}

\begin{document}

\maketitle
\thispagestyle{empty}
\pagestyle{empty}

%%%%%%%%%%%%%%%%%%%%%%%%%%%%%%%%%%%%%%%%%%%%%%%%%%%%%%%%%%%%%%%%%%%%%%%%%%%%%%%%
\begin{abstract}
Tactile sensing is particularly valuable for contact-rich robotic manipulation. Recent work has made substantial progress in tactile representation learning for robotic manipulation. However, similar tactile observations can arise from interaction conditions with very different task-risk implications, such as sensor noise, task-necessary variations, and emerging undesirable contact. Without context-grounded risk information, these cases can be ambiguous to downstream policies, leading to unnecessary corrections to benign variations or delayed responses to genuinely risky contact. To address this limitation, we propose Risk-Aware Tactile Encoding (RATE), which learns tactile representations that encode task-conditioned interaction risk. Specifically, history-conditioned prediction captures interaction context, while alert supervision associates this context with task-conditioned risk. The learned risk-aware representation complements conventional tactile features through a lightweight residual adapter. Experiments in both simulation and the real world demonstrate substantial improvements in task success, with controlled ablations confirming the complementary benefits of alert-guided learning and predictive temporal modeling.

\end{abstract}
\section{Introduction}
Humans naturally combine visual and tactile information during contact-rich manipulation. Vision provides object- and scene-level information, while touch provides local information about contact, force, friction, and slip~\cite{johansson2009tactile}. Recent visuo-tactile research has demonstrated the value of tactile feedback for manipulation, particularly in tasks such as insertion and extraction where subtle interaction states are difficult to infer from vision alone~\cite{george2025vital,zhang2025kinedex,chen2026univtac}. At the same time, advances in tactile representation learning, temporal modeling, and contact-aware manipulation have further expanded how tactile information can be learned and used in robotic manipulation~\cite{george2025vital,chen2026univtac,zhu2022contactsafety,wei2024safediff,calandra2017feeling,xue2025reactivediffusion,xu2026contactgrounded,heng2025vitacformer,niu2026trex,zheng2026omnivta,neoteai2026n0vtla},~\cite{zhang2020contactevents}.

% Declare the single-column top float early enough for the first page's right column.
\begin{figure}[!t]
    \centering
    \includegraphics[width=\columnwidth]{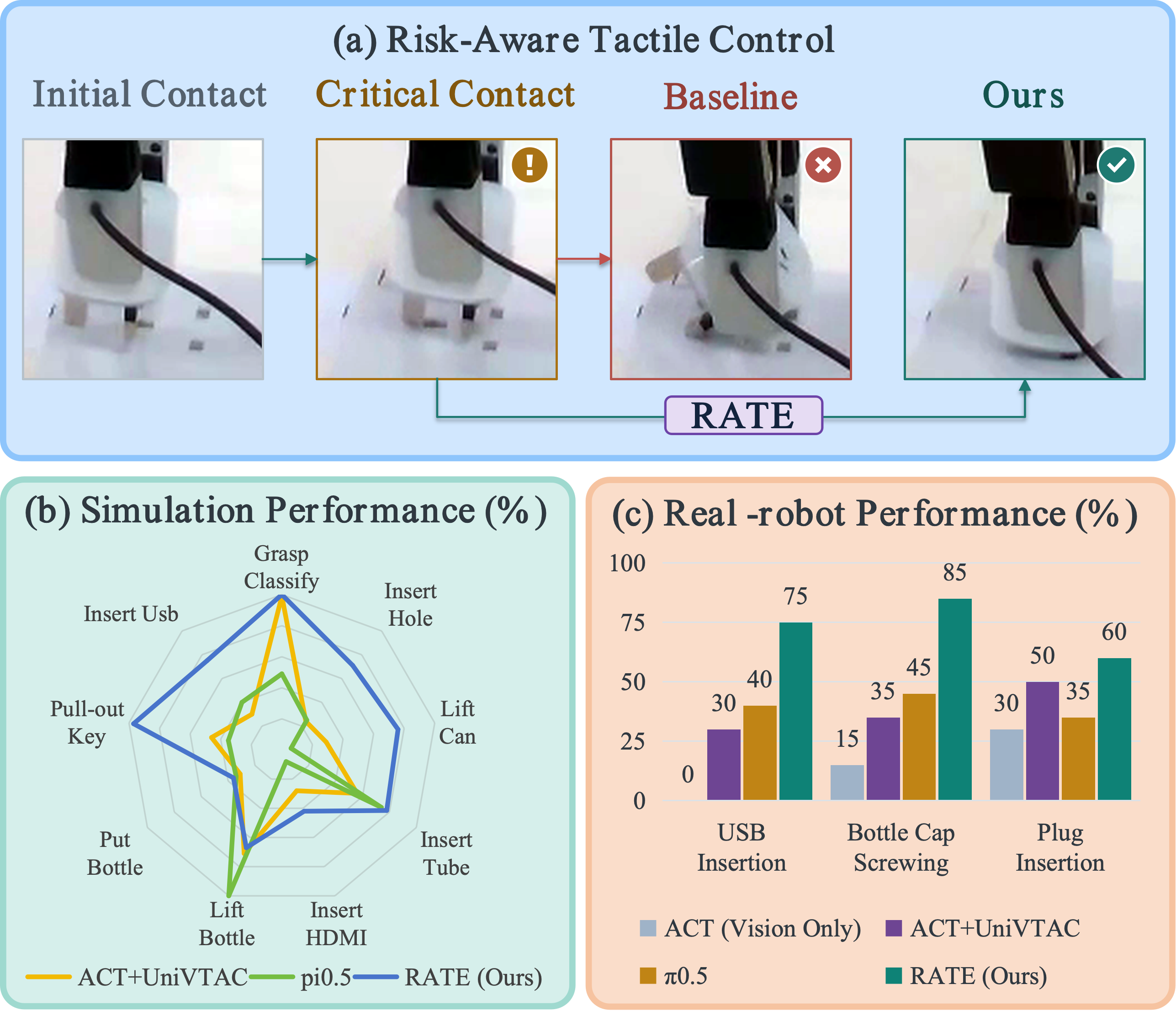}
    \caption{RATE in contact-rich manipulation: corrective behavior and task performance. (a) Representative interaction sequence: when the interaction reaches critical contact, baseline methods may continue toward failure, whereas RATE enables a corrective response toward successful execution. (b) Success rates across all eight simulation tasks, comparing RATE with representative tactile and large-scale policy baselines. RATE achieves the best or jointly best performance on six of the eight tasks. (c) Success rates on three real-robot tasks, including USB insertion, bottle-cap screwing, and plug insertion, where RATE achieves the highest success rate on all three tasks.}
    \label{fig:introduction_motivation}
\end{figure}

However, despite these advances, existing tactile representations still provide limited context-grounded information about task risk. In contact-rich manipulation, sensor noise, task-necessary adjustments, and early signs of undesirable contact can produce similar tactile observations despite corresponding to very different levels of task risk. If such risk information is not adequately captured in the representation, downstream policies may lack sufficient evidence to determine when corrective action is needed, leading to unnecessary corrections or responses that come too late.

To address this limitation, we propose \textbf{Risk-Aware Tactile Encoding (RATE)}, a tactile representation learning framework that augments conventional tactile representations with a complementary risk-aware representation. History-conditioned predictive pretraining provides contextual cues about interaction evolution by learning from tactile history and future prediction targets, while alert supervision associates these cues with task-conditioned risk. During policy adaptation, the learned risk-aware representation complements conventional tactile features through a zero-initialized residual pathway, while alert-weighted action imitation emphasizes task-critical steps. Only the residual pathway is updated, with the pretrained tactile encoders and policy backbone kept frozen, resulting in updates to approximately \textbf{0.59\%} of the complete model parameters. Alert targets and future observations are used only during training and are not required during deployment. Figure~\ref{fig:introduction_motivation} provides a qualitative and quantitative overview of RATE, illustrating its corrective behavior at critical contact and its performance across simulation and real-robot tasks.

The main contributions of this work are summarized as follows:
\begin{itemize}[topsep=0pt]
    \item \textbf{Risk-aware tactile representation learning.} We learn a risk-aware tactile representation by capturing the context of interaction evolution and associating it with task-conditioned risk through alert supervision.
    \item \textbf{Lightweight residual adaptation.} RATE integrates the learned risk-aware representation with conventional tactile features through a zero-initialized residual pathway, updating only \textbf{0.59\%} of the model parameters.
    \item \textbf{Systematic evaluation.} Simulation and real-world experiments demonstrate consistent gains, while ablations show that alert-guided learning and context are individually effective and complementary within RATE.
\end{itemize}
\section{Related Work}

In this section, we review two lines of prior work. Section~II-A focuses on visuo-tactile representation learning for contact-rich manipulation, while Section~II-B discusses temporal tactile modeling and contact-aware manipulation.

\subsection{Visuo-Tactile Representation Learning for Contact-Rich Manipulation}
Recent robotic manipulation research has advanced both dexterous task planning and motion control \cite{liu2025robodexvlm} and general-purpose gripper-based manipulation \cite{liu2026unimanip}. Visuo-tactile representation learning has evolved from simple multimodal fusion toward more structured and transferable representations. One line of work focuses on cross-modal learning, using contrastive objectives or shared latent spaces to capture complementary information between vision and touch~\cite{george2025vital,wu2025convitac}. Another line introduces stronger structural priors, such as shared 3D representations and implicit contact geometry, to preserve spatial relationships among the robot, object, and contact regions~\cite{huang2024vitac,lee2025vitascope}. In parallel, large-scale simulation and tactile-rich demonstrations have improved the scalability of representation learning and its transfer to downstream manipulation tasks~\cite{chen2026univtac,zhu2025touch}.

Together, these approaches substantially improve the representation of contact appearance, geometry, and multimodal structure. However, their learning objectives are not explicitly designed to organize tactile information according to task-conditioned risk in the context of an evolving interaction.

\subsection{Temporal Tactile Modeling and Contact-Aware Manipulation}
Recent work has extended tactile modeling beyond instantaneous observations by incorporating temporal information. One class of methods uses recurrent or spatiotemporal models to capture contact evolution from tactile sequences and predict contact events~\cite{zhang2020contactevents}. Another class adopts predictive representation learning, conditioning on current or recently observed tactile information to predict future tactile observations or tactile changes, thereby encouraging the learned representation to capture how interaction evolves over time~\cite{heng2025vitacformer,neoteai2026n0vtla}. These studies demonstrate that tactile history and future-prediction objectives provide useful temporal information for modeling interaction dynamics.

Tactile and force feedback has also been used for real-time adaptation in dexterous teleoperation \cite{liu2026dexteleop0}. Reactive approaches use high-frequency tactile feedback for online action correction~\cite{xue2025reactivediffusion,niu2026trex}, while predictive approaches model future robot--contact trajectories or visuo-tactile dynamics to guide action generation and closed-loop refinement~\cite{xu2026contactgrounded,zheng2026omnivta}. These methods improve policy responsiveness to changing contact conditions, but temporal information is primarily used for dynamics prediction or action correction.

A complementary line of work focuses on contact-aware manipulation, where tactile or contact feedback is used to assess or constrain undesirable interactions. Existing approaches consider grasp-outcome prediction, collision detection, contact-force regulation, and trajectory refinement~\cite{zhu2022contactsafety,wei2024safediff,calandra2017feeling}. These studies demonstrate the value of risk-related contact information for manipulation decisions, but such signals are typically used at the level of event prediction, control constraints, or policy optimization rather than to directly shape the tactile representation itself.

\section{Methodology}
\label{sec:method}

\begin{figure*}[!t]
    \centering
    \includegraphics[width=0.96\textwidth]{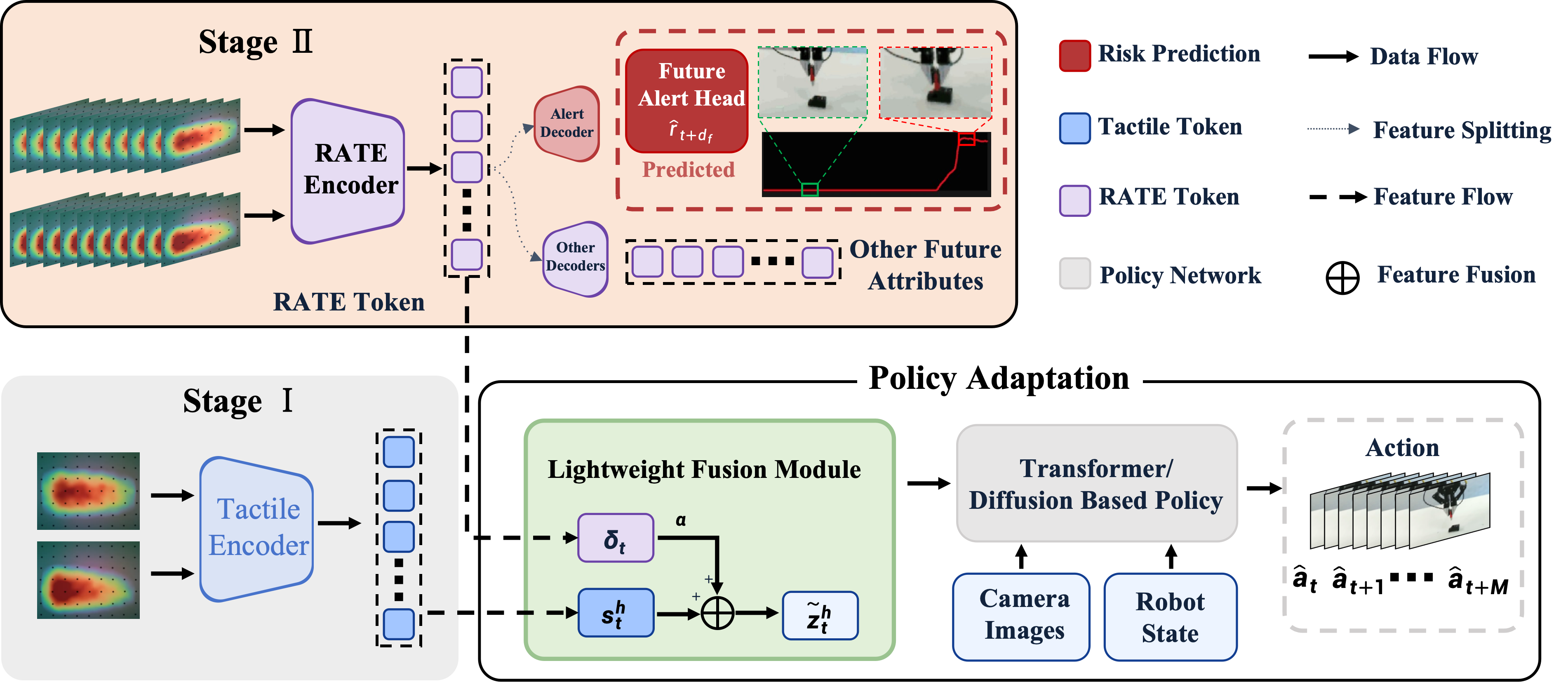}
    \caption{Overview of RATE representation learning and policy adaptation. RATE learns a complementary risk-aware representation from tactile history, jointly shaped by short-horizon future contact-change prediction and future alert supervision. During policy adaptation, the learned representation is mapped to a bounded residual and fused with conventional tactile features through a lightweight residual pathway. The resulting tactile representation, together with camera observations and robot state, conditions the downstream manipulation policy. The tactile encoders and policy backbone remain frozen during adaptation, and all auxiliary prediction heads are used only for training.}
    \label{fig:rate_pipeline}
\end{figure*}

This section presents the proposed \textbf{Risk-Aware Tactile Encoding (RATE)} framework. Section~III-A formulates the problem and outlines the overall framework. Section~III-B describes the learning of the risk-aware tactile representation, and Section~III-C details its lightweight residual integration into a pretrained manipulation policy.

\subsection{Problem Formulation and Framework Overview}

Reliable contact-rich manipulation requires tactile representations that explicitly characterize task-conditioned risk, while existing tactile representations remain limited in this regard. We therefore formulate the problem as learning a complementary risk-aware tactile representation $z_t^r$ from recent tactile interaction, using only observations available up to time $t$ during deployment. Rather than replacing the existing tactile representation, $z_t^r$ augments it with risk-related information, enabling lightweight adaptation of a pretrained manipulation policy.
% This representation is designed to complement, rather than replace, the existing tactile representation, allowing risk-related information to be introduced into a pretrained manipulation policy with lightweight adaptation.

Figure~\ref{fig:rate_pipeline} summarizes the overall RATE representation-learning and policy-adaptation pipeline. RATE learns a complementary risk-aware tactile representation grounded in interaction context and integrates it with conventional tactile features through a lightweight residual pathway. To learn this representation, RATE uses tactile history through short-horizon future prediction together with continuous task-conditioned alert supervision $r_t\in[0,1]$. Alert targets are automatically derived from task-relevant interaction signals available in each domain and normalized to a common $[0,1]$ scale. Higher alert values correspond to increasingly undesirable contact conditions. Through this joint supervision, the learned representation is encouraged to distinguish sensor noise and task-necessary variations from emerging undesirable contact conditions.

The learned risk-aware representation complements conventional tactile features through a lightweight zero-initialized residual pathway, without replacing the original tactile encoder. Future observations and alert targets are used only during training. At deployment, the policy relies solely on visual, tactile, and proprioceptive observations available up to the current time step.

\subsection{Learning Risk-Aware Tactile Representations}

In contact-rich manipulation, task risk depends on both the current contact state and how that state evolves over time. RATE therefore constructs interaction context from the bilateral contact configuration, local temporal changes, and their accumulation over a short tactile history, while continuous task-conditioned alert supervision simultaneously organizes these evolving cues according to their relevance to task risk. This allows interaction states with similar tactile appearance but different risk implications to become distinguishable in the learned representation. As shown in Stage II of Fig.~\ref{fig:rate_pipeline}, these complementary cues are jointly encoded into the risk-aware representation $z_t^r$.

At each sampled time step $k$, the left and right tactile observations are processed by a shared tactile backbone to obtain $f_k^L=E_r(\tau_k^L)$ and $f_k^R=E_r(\tau_k^R)$, which are then fused as $e_k=\psi([f_k^L,f_k^R])$. Because task risk arises from the overall contact interaction rather than either tactile sensor in isolation, $e_k$ provides a joint description of the bilateral contact configuration.

The contact configuration alone does not explicitly reveal how the interaction is evolving. We therefore compute the local feature transition $\Delta e_k=e_k-e_{k-1}$ and jointly encode the contact state and its temporal change as $u_k=\phi([e_k,\Delta e_k])$. Here, $e_k$ specifies the contact configuration in which a change occurs, while $\Delta e_k$ captures the direction and magnitude of the local transition in feature space. Their combination is important because similar local changes can have different implications under different contact configurations, while similar contact states can evolve toward different outcomes depending on how they are changing.

Risk-relevant interaction patterns may persist or accumulate across multiple local transitions rather than appearing in a single adjacent-frame difference. We therefore aggregate the sequence $\{u_k\}_{k=1}^{H}$ over the recent interaction history, so that the resulting representation captures both the evolving contact state and the temporal progression of the interaction. In our implementation, this temporal aggregation is performed using a two-layer LSTM over a tactile history of length $H=9$ with sampling stride $s=5$. The final hidden state, $z_t^r=h_H^{(2)}\in\mathbb{R}^{512}$, summarizes the recent interaction context and, under the joint predictive and alert supervision described below, serves as the risk-aware tactile representation.

During pretraining, a set of auxiliary prediction heads is attached to the same $z_t^r$ to jointly shape this interaction context with respect to task risk. For contact-related quantities $y_t$, including tactile RGB, depth, and contact pose, the corresponding decoders predict short-horizon future changes $\Delta y_t=y_{t+d_f}-y_t$, where $d_f=5$ frames. In parallel, an alert prediction head estimates the task-conditioned alert at the same future target time as $\hat r_{t+d_f}=H_r(z_t^r)$. Future contact-change prediction requires $z_t^r$ to preserve the context needed to characterize how the ongoing interaction is evolving, while future alert supervision organizes these evolving cues according to their relation to task risk. Their joint supervision therefore encourages the representation to retain tactile cues that emerge before task-critical interactions and to interpret them within the preceding interaction context.

As an additional validation of the learnability of the alert signal, we evaluate a current-alert predictor that estimates $r_t$ from current tactile features. It achieves an average MSE of $0.0256$ across validation episodes from all eight tasks. This current-alert predictor is used to assess the learnability of the continuous alert signal and is distinct from the history-conditioned future-alert objective used to learn $z_t^r$.

Because informative tactile changes are often confined to small spatial regions, uniform regression over the full prediction target can be dominated by nearly unchanged areas. For each image-based target, we therefore use training-set change statistics to emphasize regions containing substantial changes. Specifically, the 90th percentile $P90_y$ of the change magnitude is used to construct a significant-change mask, while the 99th percentile $P99_y$ is used to normalize the regression target:
\begin{equation}
    \begin{aligned}
        M_y
        &=\mathbf{1}\left(
        \frac{\operatorname{mean}_c|\Delta y_t|}{P90_y}>1
        \right),\\
        \bar{\Delta y}_t
        &=\operatorname{clip}\left(
        \frac{\Delta y_t}{P99_y},-1,1
        \right).
    \end{aligned}
\end{equation}

For each image modality, we compute a global regression loss $\mathcal{L}_{\mathrm{global}}$ over the full target and a change-focused loss $\mathcal{L}_{\mathrm{change}}$ over the significant-change regions, and combine them as
\begin{equation}
    \mathcal{L}_y
    =(1-\gamma_y)\mathcal{L}_{\mathrm{global}}
    +\gamma_y\mathcal{L}_{\mathrm{change}},
\end{equation}
where $\gamma_y$ controls the relative emphasis on regions containing substantial changes. The alert prediction head is supervised by the future target $r_{t+d_f}$ using mean squared error, yielding $\mathcal{L}_{\mathrm{risk}}$.

The complete RATE pretraining objective is
\begin{equation}
    \begin{aligned}
        \mathcal{L}_{\mathrm{RATE}}
        ={}&\lambda_{\mathrm{rgb}}\mathcal{L}_{\mathrm{rgb}}
        +\lambda_{\mathrm{depth}}\mathcal{L}_{\mathrm{depth}}\\
        &+\lambda_{\mathrm{pose}}\mathcal{L}_{\mathrm{pose}}
        +\lambda_{\mathrm{risk}}\mathcal{L}_{\mathrm{risk}}.
    \end{aligned}
\end{equation}

Here, $\gamma_y$ balances global and change-focused supervision within each image modality, whereas the $\lambda$ coefficients balance the different RGB, depth, contact-pose, and alert prediction targets. Through this joint objective, the contact configuration, local temporal transitions, and their accumulation over recent tactile history are all organized around their relevance to task risk. RATE therefore learns a risk-aware tactile representation in which evolving contact cues are interpreted according to their relevance to task risk.

Future tactile observations and alert targets are used only to construct supervision targets during pretraining. After pretraining, all auxiliary prediction heads are removed. At deployment, $z_t^r$ is computed solely from current and historical tactile observations, without access to future observations or alert targets.

\subsection{Lightweight Residual Integration into the Manipulation Policy}
\label{subsec:rate_integration}

As illustrated in the policy-adaptation stage of Fig.~\ref{fig:rate_pipeline}, the learned risk-aware representation is integrated into the pretrained manipulation policy as a lightweight residual correction to conventional tactile features. Rather than replacing the original tactile representation, this design preserves the contact information already captured by the pretrained tactile encoder and introduces the additional context-grounded risk information learned by RATE.

For each tactile sensor $h\in\{L,R\}$, let $z_t^{c,h}$ denote the feature produced by the pretrained conventional tactile encoder, and let $s_t^h=P_c(z_t^{c,h})$ be its projected tactile token. The risk-aware representation $z_t^r$, obtained from the bilateral tactile history in Sec.~III-B, is mapped into the same feature space through a lightweight residual adapter:
\begin{equation}
    \delta_t=\tanh\!\left(A(P_r z_t^r)\right),
\end{equation}
where $P_r$ is an input projection and $A(\cdot)$ is a two-layer MLP. The resulting residual is fused with the conventional tactile tokens as
\begin{equation}
    \tilde z_t^h=s_t^h+\alpha_{\mathrm{task}}\delta_t,
    \qquad h\in\{L,R\},
\end{equation}
where $\alpha_{\mathrm{task}}$ is a fixed task-level scaling coefficient. Because $z_t^r$ is encoded jointly from the bilateral tactile history, the same risk-aware residual is applied to both tactile tokens. In this way, RATE does not form a separate policy input stream. Instead, it directly complements the conventional tactile representation with interaction-context information organized around task risk.

To preserve the behavior of the pretrained policy at the beginning of adaptation, the final linear layer of $A(\cdot)$ is initialized with zero weights and bias. The residual therefore starts from $\delta_t=0$, so the fused representation initially reproduces the original tactile tokens exactly. During policy adaptation, the conventional tactile encoder, the RATE encoder, the visual backbone, and the ACT backbone remain frozen, while only $P_r$ and the residual adapter $A(\cdot)$ are optimized. This updates approximately $0.59\%$ of the complete model parameters.

The fused tactile tokens are combined with the visual observations $I_t$ and proprioceptive state $q_t$, and the pretrained ACT policy~\cite{zhao2023act} predicts a $Q=50$-step action chunk as $\hat{\mathcal A}_t=\pi_{\mathrm{ACT}}(I_t,q_t,\tilde z_t^L,\tilde z_t^R)$. To further align the residual adaptation with task risk, alert is used only during training to reweight the imitation loss across action steps. For action step $j$, the corresponding alert value determines a normalized weight $\tilde w_{b,j}$, assigning greater emphasis to steps associated with more undesirable contact conditions. The resulting risk-aware imitation objective is
\begin{equation}
    \mathcal{L}_{\mathrm{RA}}
    =\frac{1}{BQD}
    \sum_{b=1}^{B}\sum_{j=1}^{Q}
    m_{b,j}\tilde w_{b,j}
    \left\lVert a_{b,j}-\hat a_{b,j}\right\rVert_1,
\end{equation}
where $m_{b,j}$ masks padded steps and $D$ is the action dimension. The complete policy objective retains the original ACT KL regularization, giving $\mathcal{L}_{\mathrm{policy}}=\mathcal{L}_{\mathrm{RA}}+\beta_{\mathrm{KL}}\mathcal{L}_{\mathrm{KL}}$.

During adaptation, the learned risk-aware representation provides the residual correction to the conventional tactile features. Alert targets additionally reweight the imitation objective to emphasize task-critical steps. Neither alert targets nor predicted alert values are provided as policy inputs. At deployment, the policy operates only on visual observations, proprioception, and tactile observations available up to the current time step.

% Queue the two full-width experimental tables for consecutive page tops.
\begin{table*}[t]
    \centering
    \caption{Simulation success rates (\%) on the eight-task UniVTAC benchmark. Best results are bold. Published ACT, ACT+UniVTAC, and VITaL success rates are reported in UniVTAC~\cite{chen2026univtac}. The success rates of $\pi_{0.5}$, StarVLA-$\alpha$, and GigaWorld-Policy are reported in $N_0$-VTLA~\cite{neoteai2026n0vtla}. Published results are included for reference and may differ in demonstration count and observation configuration. ACT+UniVTAC (Retrained) is retrained and evaluated by us under the same configuration as RATE.}
    \label{tab:univtac_main_results}
    \small
    \setlength{\tabcolsep}{3pt}
    \renewcommand{\arraystretch}{1.1}
    \begin{tabular*}{\textwidth}{@{\extracolsep{\fill}}l*{10}{c}@{}}
        \toprule
        \multirow{2}{*}{Method}
        & \multirow{2}{*}{Params.}
        & \multicolumn{4}{c}{Contact-Rich Interaction}
        & \multicolumn{3}{c}{Pose Reasoning}
        & \multicolumn{1}{c}{\makecell{Shape\\Perception}}
        & \multirow{2}{*}{Avg.} \\
        \cmidrule(lr){3-6}\cmidrule(lr){7-9}\cmidrule(lr){10-10}
        & & \makecell{Pull-out\\Key}
        & \makecell{Insert\\Hole}
        & \makecell{Insert\\HDMI}
        & \makecell{Insert\\Tube}
        & \makecell{Lift\\Bottle}
        & \makecell{Lift\\Can}
        & \makecell{Put\\Bottle}
        & \makecell{Grasp\\Classify}
        & \\
        \midrule
        ACT (Vision Only)~\cite{zhao2023act} & 0.08B & 28.0 & 19.0 & 15.0 & 45.0 & 42.0 & 20.0 & 28.0 & 50.0 & 30.9 \\
        ACT+UniVTAC~\cite{chen2026univtac} & 0.10B & 46.0 & 24.0 & 28.0 & 56.0 & 71.0 & 29.0 & 31.0 & 99.0 & 48.0 \\
        VITaL~\cite{george2025vital} & 0.10B & 47.0 & 25.0 & 6.0 & 34.0 & 72.0 & 8.0 & 32.0 & \textbf{100.0} & 40.5 \\
        $\pi_{0.5}$~\cite{physicalintelligence2025pi05} & 3.3B & 35.0 & 25.0 & 8.0 & 74.0 & \textbf{100.0} & 6.0 & 34.0 & 49.0 & 41.4 \\
        StarVLA-$\alpha$~\cite{ye2026starvlaalpha} & 4B & 51.0 & 52.0 & 24.0 & 69.0 & 32.0 & 65.0 & \textbf{88.0} & 68.0 & 56.1 \\
        GigaWorld-Policy~\cite{ye2026gigaworldpolicy} & 5B & 32.0 & 12.0 & 0.0 & 9.0 & 38.0 & 0.0 & 21.0 & 20.0 & 16.5 \\
        ACT+UniVTAC (Retrained) & 0.10B & 59.0 & 23.0 & 15.0 & 37.0 & 89.0 & 47.0 & 10.0 & 99.0 & 47.4 \\
        \midrule
        \textbf{RATE (Ours)} & 0.133B & \textbf{97.0} & \textbf{71.0} & \textbf{42.0} & \textbf{78.0} & 74.0 & \textbf{76.0} & 36.0 & \textbf{100.0} & \textbf{71.8} \\
        \bottomrule
    \end{tabular*}
\end{table*}

\begin{table*}[t]
    \centering
    \caption{Ablation results for RATE on the eight-task UniVTAC benchmark. All variants share the same task-specific tactile base. The table evaluates alert supervision using Discrete Ordinal, Continuous Severity, and Transition-Aware Continuous formulations, while context denotes interaction context learned through history-conditioned prediction. The final row corresponds to Full RATE, combining continuous severity alert supervision with context. Avg. denotes the eight-task macro average. Best results are bold.}
    \label{tab:ablation_results}
    \small
    \setlength{\tabcolsep}{1.5pt}
    \renewcommand{\arraystretch}{1.1}
    \begin{tabularx}{\linewidth}{@{}>{\centering\arraybackslash}m{3.8em}>{\centering\arraybackslash}m{5.0em}>{\centering\arraybackslash}m{4.8em}>{\centering\arraybackslash}m{3.8em}@{\hspace{7pt}}*{9}{Y}@{}}
        \toprule
        \multicolumn{3}{c}{Alert Supervision}
        & \multirow{2}{*}{\raisebox{-1.2ex}{Context}}
        & \multicolumn{4}{c}{Contact-Rich Interaction}
        & \multicolumn{3}{c}{Pose Reasoning}
        & \multicolumn{1}{c}{\makecell{Shape\\Perception}}
        & \multirow{2}{*}{Avg.} \\
        \cmidrule(lr){1-3}\cmidrule(lr){5-8}\cmidrule(lr){9-11}\cmidrule(lr){12-12}
        Discrete
        & Continuous
        & \makecell{Transition-\\Aware}
        &
        & \makecell{Pull-out\\Key}
        & \makecell{Insert\\Hole}
        & \makecell{Insert\\HDMI}
        & \makecell{Insert\\Tube}
        & \makecell{Lift\\Bottle}
        & \makecell{Lift\\Can}
        & \makecell{Put\\Bottle}
        & \makecell{Grasp\\Classify}
        & \\
        \midrule
        -- & -- & -- & -- & 51.0 & 30.0 & 16.0 & 49.0 & 51.0 & 49.0 & 7.0 & 99.0 & 44.0 \\
        $\checkmark$ & -- & -- & -- & 65.0 & 31.0 & 31.0 & 59.0 & 67.0 & 62.0 & 34.0 & \textbf{100.0} & 56.1 \\
        -- & $\checkmark$ & -- & -- & 81.0 & 32.0 & 18.0 & 66.0 & 61.0 & 67.0 & 35.0 & \textbf{100.0} & 57.5 \\
        -- & -- & $\checkmark$ & -- & 78.0 & 41.0 & 14.0 & 66.0 & 68.0 & 52.0 & 29.0 & \textbf{100.0} & 56.0 \\
        -- & -- & -- & $\checkmark$ & 68.0 & 46.0 & 25.0 & 56.0 & 73.0 & 68.0 & 12.0 & \textbf{100.0} & 56.0 \\
        \midrule
        -- & $\boldsymbol{\checkmark}$ & -- & $\boldsymbol{\checkmark}$ & \textbf{97.0} & \textbf{71.0} & \textbf{42.0} & \textbf{78.0} & \textbf{74.0} & \textbf{76.0} & \textbf{36.0} & \textbf{100.0} & \textbf{71.8} \\
        \bottomrule
    \end{tabularx}
\end{table*}

\section{Experiments}

We evaluate RATE on eight simulated manipulation tasks and three real-world contact-rich tasks. Our experiments address three questions: (1) Can RATE outperform representative manipulation policies overall without relying on large-scale models? (2) How do alert-guided learning and interaction context contribute, individually and jointly, to RATE, and how effectively can the resulting risk-aware representation be integrated through lightweight residual adaptation? (3) How effectively does RATE achieve both task completion and contact-safe execution in real-world settings?

\subsection{Experimental Setup}
\label{subsec:experimental_setup}

\textit{Simulation benchmark.} We evaluate RATE on the eight simulated tasks of the UniVTAC benchmark~\cite{chen2026univtac}. Following the official taxonomy, Contact-Rich Interaction includes Pull-out Key, Insert Hole, Insert HDMI, and Insert Tube. Pose Reasoning includes Lift Bottle, Lift Can, and Put Bottle. Shape Perception includes Grasp Classify. All tasks use a Franka Panda robot equipped with a parallel gripper and two simulated GelSight Mini tactile sensors. The policy receives RGB images, bilateral marker-based tactile images, and an eight-dimensional proprioceptive state.

In our internal experiments, we use 100 demonstrations per task, with head- and wrist-camera RGB inputs across all eight tasks. Visual and tactile observations are resized to $256\times256$, and robot states and actions are normalized using statistics from the corresponding training set. We use ACT's action chunking and temporal aggregation~\cite{zhao2023act}, with a 50-step action chunk in our configuration. We compare RATE with ACT (Vision Only)~\cite{zhao2023act}, ACT+UniVTAC~\cite{chen2026univtac}, VITaL~\cite{george2025vital}, $\pi_{0.5}$~\cite{physicalintelligence2025pi05}, StarVLA-$\alpha$~\cite{ye2026starvlaalpha}, and GigaWorld-Policy~\cite{ye2026gigaworldpolicy}. We report published success rates for these baselines~\cite{chen2026univtac,neoteai2026n0vtla}. For a fair comparison, we additionally retrain ACT+UniVTAC using the same demonstrations and camera configuration as RATE. RATE, ACT+UniVTAC (Retrained), and our internally trained variants are evaluated under the same protocol over 100 closed-loop rollouts per task. We report task-wise success rates, the eight-task macro average, and model scale.

\textit{Ablation protocol.} All ablation variants share the same task-specific static tactile base, trained separately for each task using the same task-specific demonstrations as RATE and the UniVTAC-style tactile representation learning procedure. Starting from this common task-specific tactile base, we vary two factors: alert and context. The alert is evaluated using three formulations---Discrete Ordinal, Continuous Severity, and Transition-Aware Continuous---while context is introduced through history-conditioned predictive modeling. The task-specific tactile base contains neither factor; the alert-only configurations enable one alert formulation without context; the context-only configuration introduces context without alert-related training; and Full RATE combines the selected continuous severity alert formulation with context. All other settings, including demonstrations, visual inputs, ACT backbone, action representation, data split, and evaluation protocol, are kept identical.

\textit{Real-world setup.} We further evaluate RATE on USB Insertion, Bottle Cap Screwing, and Plug Insertion. All three tasks use the same platform: an xArm 7 robot with an xArm parallel gripper, two Daimon tactile sensors, a RealSense D435 head camera, and a RealSense D405 wrist camera. We collect 50 task-specific demonstrations for each task. To retain a consistent image-based tactile interface, the native Daimon measurements are converted into GelSight-like RGB tactile images, on which the tactile encoder is retrained.

Each method is evaluated over 20 trials per task. USB Insertion and Plug Insertion are considered successful when the USB connector and plug, respectively, are fully inserted. For Bottle Cap Screwing, a trial is considered successful if tightening does not wrinkle the tablecloth and the cap cannot be unscrewed with light manual force after execution.

In addition to Task Success Rate (SR), we report Contact-Safe Success Rate (CSSR), defined as the proportion of successful trials that also satisfy the contact-safety criterion:
\begin{equation}
    \begin{aligned}
        \mathrm{SR} &= \frac{N_{\mathrm{success}}}{N_{\mathrm{total}}}, \\
        \mathrm{CSSR} &= \frac{N_{\mathrm{safe}}}{N_{\mathrm{success}}},
    \end{aligned}
\end{equation}
where $N_{\mathrm{total}}$ and $N_{\mathrm{success}}$ denote the total number of trials and the number of successful trials, respectively. $N_{\mathrm{safe}}$ counts successful trials in which the maximum estimated tactile depth remains below a predefined threshold throughout execution. The threshold is fixed before evaluation and applied consistently across methods. Maximum tactile depth serves as a proxy for excessive tactile deformation rather than a direct measurement of contact force. For reporting, CSSR is set to 0 when no trials succeed.

\subsection{Simulation Results on the UniVTAC Benchmark}
\label{subsec:simulation_results}

As shown in Table~\ref{tab:univtac_main_results}, RATE achieves the highest macro-average success rate of 71.8\% across the eight tasks, ranking first or jointly first on six. It leads all four Contact-Rich Interaction tasks, achieves the highest category-average performance in Pose Reasoning, and ties for the best result in Shape Perception. Notably, RATE reaches an average success rate of 72.0\% across the four Contact-Rich Interaction tasks, exceeding the strongest comparison method in this category, StarVLA-$\alpha$, by 23.0 percentage points. This pattern is consistent with RATE's design goal of representing task risk in the context of evolving tactile interaction.

RATE also maintains a clear advantage under the controlled experimental setting. When ACT+UniVTAC is retrained using the same demonstrations and camera configuration as RATE, RATE achieves a macro-average success rate of 71.8\%, compared with 47.4\% for ACT+UniVTAC (Retrained). On the four Contact-Rich Interaction tasks, RATE improves success rates by 38, 48, 27, and 41 percentage points on Pull-out Key, Insert Hole, Insert HDMI, and Insert Tube, respectively. For a fairer comparison with UniVTAC, we additionally retrain ACT+UniVTAC using the same demonstrations and camera configuration as RATE. This provides a same-configuration external comparison, while RATE's controlled ablations are evaluated relative to the task-specific tactile base in Sec.~\ref{subsec:ablation_studies}.

Beyond Contact-Rich Interaction, RATE also achieves leading performance in the other task categories. It attains the highest category-average success rate of 62.0\% on Pose Reasoning and ties for the best result in Shape Perception with 100\% success on Grasp Classify. These results demonstrate strong performance across task categories, with the most pronounced gains on the contact-rich interactions targeted by RATE.

With approximately 133.2M parameters, RATE achieves a higher reported macro-average success rate than all compared methods, including large-scale visuomotor policies with 3.3B--5B parameters. In particular, RATE exceeds the strongest large-scale baseline, StarVLA-$\alpha$, by 15.6 percentage points based on unrounded averages.

Overall, RATE achieves the highest reported macro-average success rate across the benchmark and maintains leading performance across all three task categories, despite using a substantially smaller model than the large-scale visuomotor policies considered. These results demonstrate that strong overall manipulation performance can be achieved without relying on large-scale models.

\subsection{Ablation Studies}
\label{subsec:ablation_studies}

Table~\ref{tab:ablation_results} reports the controlled ablation results for alert-guided learning and context under the protocol defined in Sec.~\ref{subsec:experimental_setup}. Starting from the common task-specific tactile base, we compare three alert formulations, context alone, and their combination in Full RATE.

The task-specific tactile base achieves a macro-average success rate of 44.0\%. Adding the three alert formulations increases the macro average to 56.1\%, 57.5\%, and 56.0\% for Discrete Ordinal, Continuous Severity, and Transition-Aware Continuous, respectively. Continuous Severity Risk achieves the best overall result and is therefore adopted in Full RATE. Independently introducing context yields a macro-average success rate of 56.0\%, corresponding to a 12.0 percentage-point improvement over the task-specific tactile base.

When continuous alert and context are jointly incorporated in Full RATE, the macro-average success rate further increases to 71.8\%, outperforming the task-specific tactile base, continuous alert configuration, and context-only configuration by 27.8, 14.3, and 15.8 percentage points, respectively.

We further use the continuous alert configuration as the direct controlled baseline for residual adaptation to evaluate the additional benefit of integrating RATE's learned risk-aware representation. On the Contact-Rich Interaction tasks, the average success rate increases from 49.3\% to 72.0\%, including improvements from 32\% to 71\% on Insert Hole and from 18\% to 42\% on Insert HDMI. This comparison keeps the task-specific static tactile representation and alert-guided training fixed while additionally introducing context and the learned residual representation, thereby isolating the benefit of RATE's history-conditioned risk-aware adaptation. During this stage, the tactile encoders, visual backbone, and ACT backbone remain frozen, with only approximately 0.59\% of the model parameters updated.

Taken together, these results show that alert-guided learning and context each provide independent gains over the common task-specific tactile base, while their combination in Full RATE yields a further improvement, supporting their complementary roles within the same risk-aware representation.

\subsection{Real-World Evaluation}
\label{subsec:real_world_evaluation}

\begin{figure}[H]
    \centering
    \includegraphics[width=\columnwidth]{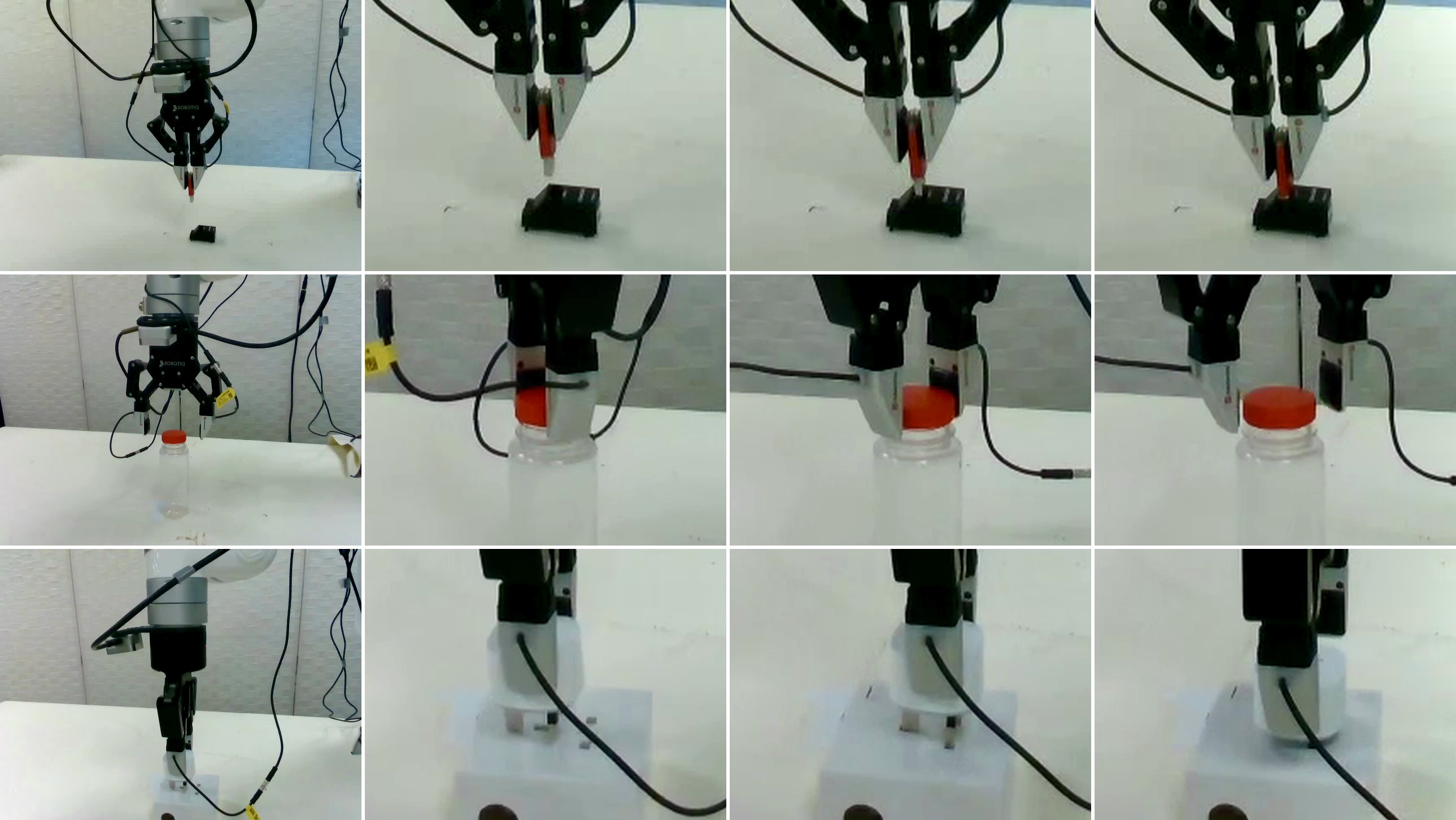}
    \caption{Real-world contact-rich manipulation tasks. Rows from top to bottom show USB Insertion, Bottle Cap Screwing, and Plug Insertion. Columns show successive stages of each task from left to right.}
    \label{fig:real_world_tasks}
\end{figure}

We evaluate RATE on USB Insertion, Bottle Cap Screwing, and Plug Insertion against ACT (Vision Only), ACT+UniVTAC, and $\pi_{0.5}$. All methods use the same 50 task-specific demonstrations and are evaluated from the same 20 initial conditions. Table~\ref{tab:real_world_results} reports SR and CSSR as defined in Sec.~\ref{subsec:experimental_setup}, while Fig.~\ref{fig:real_world_tasks} illustrates the three tasks.

% Author-provided real-world results for all three tasks.
\begin{table}[H]
    \centering
    \caption{Real-world evaluation over 20 trials per task. Task Success Rate (SR) and Contact-Safe Success Rate (CSSR) are reported in separate columns, in percent. Best results are bold.}
    \label{tab:real_world_results}
    \setlength{\tabcolsep}{2.8pt}
    \renewcommand{\arraystretch}{1.05}
    \footnotesize
    \begin{tabular*}{\columnwidth}{@{\extracolsep{\fill}}l*{6}{c}@{}}
        \toprule
        \multirow{2}{*}{Method}
        & \multicolumn{2}{c}{USB Insertion}
        & \multicolumn{2}{c}{\makecell{Bottle Cap\\Screwing}}
        & \multicolumn{2}{c}{Plug Insertion} \\
        \cmidrule(lr){2-3}\cmidrule(lr){4-5}\cmidrule(lr){6-7}
        & SR & CSSR & SR & CSSR & SR & CSSR \\
        \midrule
        ACT (Vision Only) & 0.0 & 0.00 & 15.0 & 33.33 & 30.0 & 33.33 \\
        ACT+UniVTAC       & 30.0 & 0.00 & 35.0 & 85.71 & 50.0 & 70.00 \\
        $\pi_{0.5}$       & 40.0 & 62.50 & 45.0 & 55.56 & 35.0 & 57.14 \\
        \midrule
        \textbf{RATE (Ours)}
                          & \textbf{75.0} & \textbf{93.33} & \textbf{85.0} & \textbf{100.00} & \textbf{60.0} & \textbf{83.33} \\
        \bottomrule
    \end{tabular*}
\end{table}

\textbf{Task completion.} RATE achieves success rates of 75\%, 85\%, and 60\% on USB Insertion, Bottle Cap Screwing, and Plug Insertion, respectively. On USB Insertion and Bottle Cap Screwing, it exceeds $\pi_{0.5}$, the strongest comparison method in terms of SR on both tasks, by 35 and 40 percentage points. On Plug Insertion, RATE exceeds ACT+UniVTAC, the strongest comparison method on this task, by 10 percentage points. These gains demonstrate improved task completion across connector insertion and rotational tightening tasks. Notably, the strongest comparison method differs across tasks: $\pi_{0.5}$ performs best among the comparison methods on USB Insertion and Bottle Cap Screwing, whereas ACT+UniVTAC performs best on Plug Insertion. RATE maintains a clear advantage under both comparison settings, indicating that its improvement is not limited to a single task or baseline. Across all three tasks, RATE achieves a macro-average SR of 73.3\%, compared with 40.0\% for $\pi_{0.5}$, a difference of 33.3 percentage points.

\textbf{Contact-safe execution.} The improved task completion is accompanied by a higher proportion of contact-safe successful executions across all three tasks. On USB Insertion, 14 of RATE's 15 successful trials satisfy the tactile-deformation limit, yielding a CSSR of 93.33\%, compared with 62.50\% for $\pi_{0.5}$, the best-performing comparison method under this metric. On Bottle Cap Screwing, all 17 successful RATE trials satisfy the criterion, yielding a CSSR of 100\%, compared with 85.71\% for ACT+UniVTAC, the strongest comparison method in CSSR on this task. On Plug Insertion, 10 of RATE's 12 successful trials satisfy the criterion, yielding a CSSR of 83.33\%, compared with 70.00\% for ACT+UniVTAC, the strongest comparison method under this metric. RATE therefore achieves the highest CSSR on each of the three real-world tasks, showing that its gains in task completion are accompanied by a larger fraction of successful executions satisfying the tactile-deformation criterion. This pattern holds for both connector-insertion tasks and the rotational Bottle Cap Screwing task, rather than being confined to a single interaction type. Across all three tasks, RATE achieves a macro-average CSSR of 92.22\%, compared with 58.40\% for $\pi_{0.5}$, the strongest comparison method by this aggregate metric.

Together, the results across the three real-world tasks answer the third research question by demonstrating higher task success on every task and a higher macro-average proportion of successful executions that remain within the prescribed tactile-deformation limit.

\section{Conclusion}

We presented RATE, a risk-aware tactile representation learning framework for contact-rich robotic manipulation. RATE grounds task-conditioned risk in evolving tactile interaction context through history-conditioned prediction and alert supervision, and complements conventional tactile features through a lightweight residual pathway for parameter-efficient risk-aware adaptation. Experiments across eight simulated tasks and three real-world tasks demonstrate that RATE consistently improves manipulation performance without relying on large-scale models, with particularly strong gains on contact-rich interactions. Ablation studies further confirm the complementary effects of alert-guided learning and context within RATE. Real-world evaluations additionally show improved task completion together with a higher proportion of contact-safe successful executions. Although RATE explicitly models local temporal changes through feature differences, subtle but task-relevant tactile variations may still become less distinguishable when contact magnitudes vary substantially across interaction stages. Future work will therefore explore scale-adaptive tactile encoding that preserves fine-grained relative contact variations while retaining their absolute interaction magnitude.

%%%%%%%%%%%%%%%%%%%%%%%%%%%%%%%%%%%%%%%%%%%%%%%%%%%%%%%%%%%%%%%%%%%%%%%%%%%%%%%%
\bibliographystyle{IEEEtran} 
\bibliography{ref} 

\end{document}